\documentclass[10pt,twocolumn,letterpaper]{article}

\usepackage{cvpr}              

\newcommand{\thename}[0]{RAD-2}

\newcommand{\tablestyle}[2]{\setlength{\tabcolsep}{#1}\renewcommand{\arraystretch}{#2}\centering\footnotesize}
\usepackage{pifont}
\usepackage{multirow}
\usepackage[table]{xcolor}
\usepackage{makecell}
\usepackage{bbm}
\usepackage{marvosym} 
\usepackage{array}
\usepackage{makecell}
\usepackage{colortbl}
\usepackage{sectsty}
\usepackage{enumitem}
\usepackage{graphicx}

\usepackage{float} 
\usepackage{titlesec}
\usepackage{algorithm}
\usepackage{algorithmic}
\newcommand{\algbox}[1]{%
  \footnotesize\textcolor{gray!100}{#1}%
}

\titlespacing*{\section}
{0pt}
{10pt plus 2pt minus 2pt}  
{6pt plus 2pt minus 2pt}   

\titlespacing*{\subsection}
{0pt}
{8pt plus 2pt minus 2pt}
{4pt plus 2pt minus 2pt}

\definecolor{horizonblue}{RGB}{0, 102, 204}

\definecolor{cvprblue}{rgb}{0.21,0.49,0.74}
\definecolor{ourlightblue}{RGB}{245,247,255}
\usepackage[pagebackref,breaklinks,colorlinks,allcolors=horizonblue]{hyperref}
\def\thename{Stream Forcing}

\def\paperID{*****} 
\def\confName{CVPR}
\def\confYear{2026}

\definecolor{horizonblue}{RGB}{0, 102, 204}

\title{
    \vspace{-4em} 
    \noindent
    \makebox[\textwidth][s]{ 
        \includegraphics[height=2em]{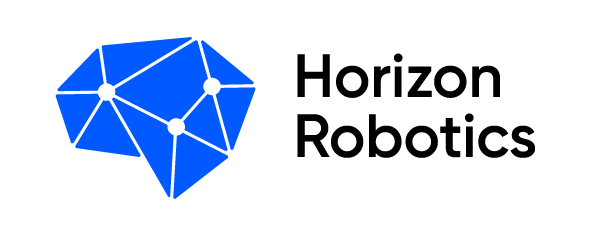} 
        \hspace{-0.5em}
        \raisebox{0.3em}{\color{gray!50}\rule{0.5pt}{1.2em}} \hspace{0.5em} 
        \includegraphics[height=2em]{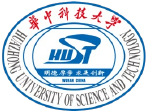} %
        \hfill 
    }
    \\
    \vspace{0.4em}
    {\color{horizonblue}\hrule height 1pt} 
    \vspace{1.0em}  
    Stream Forcing: Constructing Unified Training Trajectory \\ for Robust Streaming Video Generation
    \vspace{1.0em} \\
    {\color{horizonblue}\hrule height 1pt} 

}

\author{
\textbf{Yueting Zhu}$^{1, *}$ \quad
\textbf{Yuehao Song}$^{1, *}$ \quad
\textbf{Kaicheng Zhang}$^{3}$ \quad 
\textbf{Bao Tang}$^{1}$ \\
\textbf{Shaoyu Chen}$^{2}$ \quad  
\textbf{Qian Zhang}$^{2}$ \quad 
\textbf{Wenyu Liu}$^{1}$ \quad 
\textbf{Xinggang Wang}$^{1,\textrm{\Letter}}$ \\
\textsuperscript{1}\,Huazhong University of Science \& Technology \quad
\textsuperscript{2}\,Horizon Robotics \quad
\textsuperscript{3}\,Anyverse Dynamics \\
}

\begin{document}
\maketitle

\let\thefootnote\relax\footnotetext{$^*$Equal contribution. $^\textrm{\Letter}$ Corresponding author.}

\begin{abstract}
Streaming video generation holds strong potential for world modeling, where future frames must be inferred online sequentially to form a continuous video stream.
However, streaming video diffusion models introduce a fundamental train-inference mismatch: inference follows a specialized denoising order, whereas advanced training strategies typically require diverse noise-level configurations.
To address this trade-off between train-inference consistency and training coverage, we reformulate the video diffusion sampling as a frame-indexed stochastic process over noise levels.
Within this stochastic process space, we construct a continuous training trajectory along which the sampling schedule progressively evolves from independent sampling to inference-consistent sampling.
We further introduce a joint calibration algorithm and a temporal correlative sampling algorithm to ensure trajectory smoothness and cross-frame correlation.
Building on these designs, we propose \thename{}, a unified training framework for streaming video generation that balances training sufficiency and inference efficiency.
Extensive experiments demonstrate that \thename{} significantly improves generation quality with a 36.6\% FVD improvement on the UCF-101 benchmark.
Furthermore, our method facilitates robust zero-shot extrapolation to long-horizon video generation with a 27.9\% FVD improvement on the UCF-101 benchmark.
\end{abstract}
\vspace{-12pt}    
\section{Introduction}

Recent advances in video generation, particularly video diffusion models \cite{wan,kling,seedance}, have made remarkable progress in visual fidelity and temporal consistency.
Beyond content creation, such models can serve as generative world models \cite{worldmodel,sora} for reasoning about future observations in sequential planning, such as embodied AI \cite{chi2025empowering,worldsimbench} and autonomous driving \cite{rad,drivelaw}.
In these settings, the generative model must operate online sequentially, i.e., perform streaming inference using partial temporal context.

Unlike standard bidirectional generation \cite{latte,svd,vdm}, streaming inference requires progressively generating new frames to form a video stream in a causal setting.
Recent approaches \cite{rdm,ardiff,fifo} introduce streamlined generation by overlapping denoising across frames to reduce the high inter-frame latency.
To support this inference strategy, two training paradigms have been explored.
Progressive sampling, e.g., rolling diffusion models \cite{rdm}, improves training-inference consistency while restricting the coverage of the training distribution.
Alternatively, independent sampling, e.g., diffusion forcing \cite{df}, enables diverse training configurations while breaking the temporal structure assumed at inference time.

Noting the complementary nature of the two paradigms, we reformulate training-time noise level sampling as a frame-indexed stochastic process.
Specifically, the per-frame noise level distribution is modeled using the widely adopted Logit-normal distribution \cite{logitnorm,sd3}.
We parameterize this process with the two inherent parameters, which control temporal bias and sampling diversity, together with a temporal correlation parameter that captures dependencies between video frames.
Within this parameterized space, independent sampling is characterized by frame-invariant distribution parameters and zero temporal correlation, whereas progressive sampling exhibits frame-dependent bias and strong temporal correlation.
Taking these two configurations as endpoints, we construct a continuous training trajectory that gradually transforms the independent into the progressive configuration, thereby reconciling broad training coverage with inference consistency.

Building on this parameterization, we propose \thename{}, a unified training paradigm for progressively aligning training-time sampling with the inference schedule.
To ensure a smooth transition along the trajectory, we develop a joint calibration algorithm to heuristically determine the distribution parameters at each training step.
Meanwhile, we introduce a temporal correlative sampling algorithm to generate noise levels with the prescribed inter-frame dependencies.
These components enable training to proceed from independent per-frame sampling, through intermediate calibrated configurations, to inference-consistent sampling.

Experimental results demonstrate that \thename{} achieves high video generation quality with FVD improvements of 36.6\% on UCF-101 and 4.7\% on Taichi-HD, while also generalizing effectively to zero-shot long-horizon generation in a zero-shot setting, yielding FVD improvements of 27.9\% and 10.9\%, respectively.
We further demonstrate that \thename{} can be effectively applied to autonomous driving world modeling, improving both FID and FVD on the nuScenes \cite{nuscenes} dataset.

Our contributions can be summarized as follows:
\begin{itemize}
    \item We reformulate training-time noise-level sampling as a frame-indexed stochastic process, establishing a unified space that encompasses mainstream training paradigms.
    \item We propose \thename{}, a unified training framework that combines joint calibration and temporal correlative sampling to construct a continuous training trajectory, enabling smooth transitions throughout training while reconciling diverse coverage with inference consistency.
    \item Extensive experiments demonstrate that \thename{} achieves superior generation quality and generalizes effectively to long-horizon generation and world modeling.
\end{itemize}
\section{Background}
\subsection{Inference Strategy for Streaming Video Generation}
\begin{figure}[t]
  \centering
  \includegraphics[width=\columnwidth]{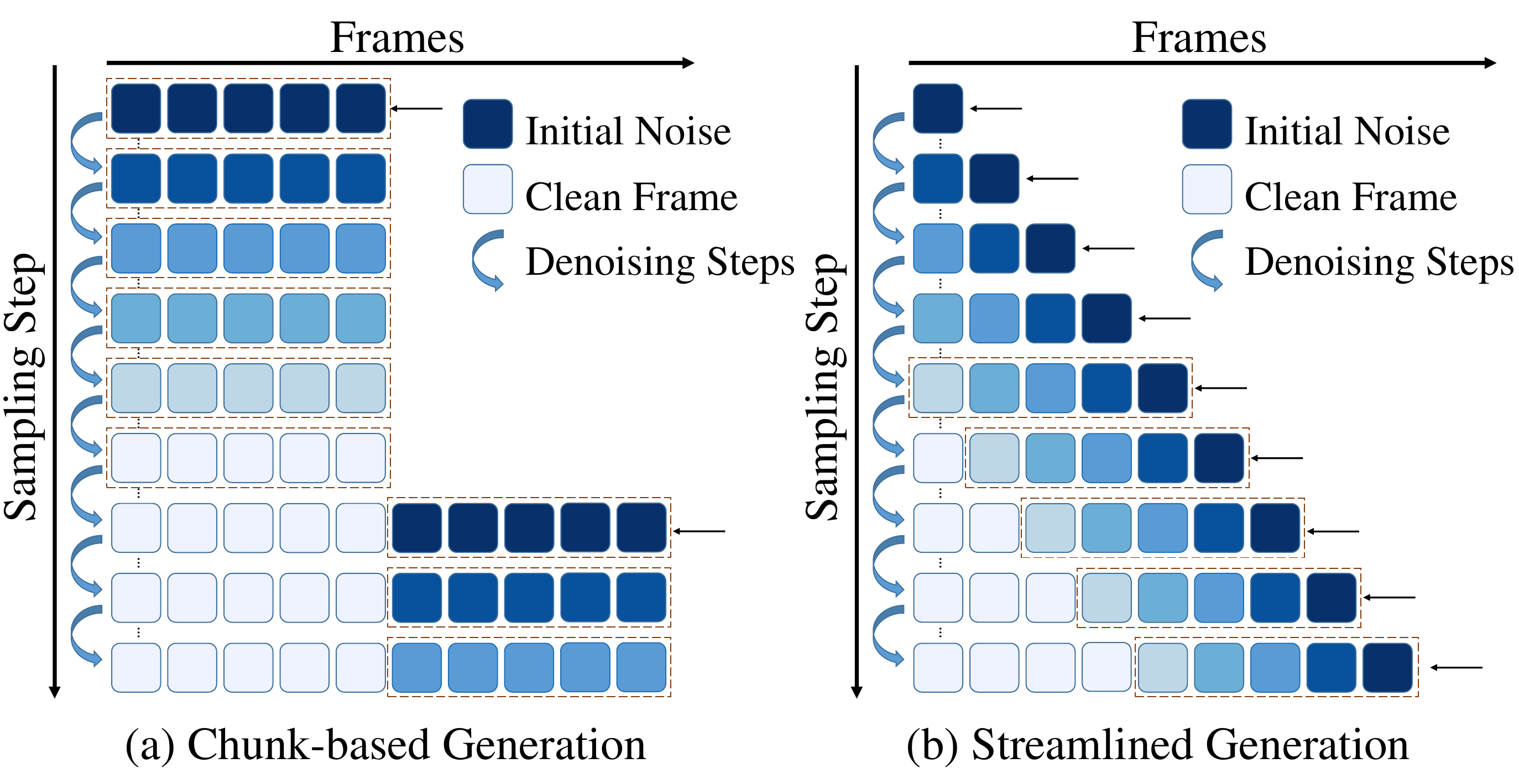}
  \caption{\textbf{Comparison of streaming inference methods.}(a) Chunk-based Generation.(b) Streamlined Generation.}
  \label{fig:inference_paradigms}
\end{figure}

Existing streaming video inference strategies mainly include chunk-based generation \cite{dfot,ca2vdm} and streamlined generation \cite{fifo,ardiff}.
As shown in Fig.~\ref{fig:inference_paradigms}, for a diffusion model with $S$ denoising steps over a $T$-frame window, chunk-based methods generate each chunk sequentially, incurring an inter-frame latency of up to $S$ denoising steps.
In contrast, streamlined methods interleave denoising within the sliding window, jointly processing frames at progressive noise levels and reducing the latency to $S/T$ steps.
We adopt streamlined inference and develop a corresponding training paradigm to improve train–inference consistency.

\subsection{Trade-off in Training}
During training, the input clip $\mathbf{x} = \{x_1, x_2, \dots, x_T\}$ of $T$ frames is corrupted by Gaussian noise at frame-specific noise levels $k_t$\footnote{There are two equivalent ways to represent the noise level: using discrete denoising steps or a continuous noise level defined in the range [0,1]. In this paper, we adopt the latter.}, resulting in a noisy observation
\begin{equation}
    x_t^{(k_t)} = \sqrt{\bar{\alpha}_{k_t}} x_t + \sqrt{1 - \bar{\alpha}_{k_t}} \epsilon_t, \quad \epsilon_t \sim \mathcal{N}(0, \mathbf{I}),
\end{equation}
where $\bar{\alpha}_{k_t}$ denotes the cumulative variance.
Under this formulation, two training strategies have been explored.

Progressive sampling~\cite{rdm} enforces monotonically increasing noise levels across frames within each training window in which the noise levels $\{k_t\}_{t=1}^T$ satisfy
\begin{equation}
    k_t \le k_{t+1}, \quad \forall\; t \in \{1, \dots, T-1\}.
\end{equation}
While such a strictly increasing design improves consistency with inference, it confines training to a narrow subset of noise schedules and reduces coverage of the training distribution.

Alternatively, independent sampling \cite{df} assigns noise levels independently to frames within the training window.
Formally, the noise levels $\{k_t\}_{t=1}^T$ satisfy
\begin{equation}
    k_t \sim p(k_1), \quad \text{i.i.d. for }\; t \in \{1, \dots, T\}.
\end{equation}
This design enables diverse noise configurations in training, while independent noise levels across frames ignore the structured noise progression assumed at inference time, which leads to a mismatch between training and inference.

\section{Method}

Motivated by the trade-off in training, we reformulate per-frame noise level sampling as a frame-indexed stochastic process to unify both training strategies.
We further construct a continuous training trajectory using the joint calibration and temporal correlative sampling algorithms.
By combining them, we develop \thename{}, a unified curriculum training procedure for streaming video generation.

\subsection{Reformulation of Training Sampling}
\label{sec:3.1reformulation}
\subsubsection{Preliminary: Logit-Normal Distribution}
Logit-normal distribution \cite{logitnorm,sd3} is widely used to sample the diffusion noise level, which is formulated with a probability density function (PDF)
{\small\begin{equation}\label{eq:pdf}
f(k) = \frac{1}{\sigma \sqrt{2\pi} k(1-k)}\exp \left(-\frac{\Big(\log\big(k/(1-k)\big) - \mu\Big)^2}{2\sigma^2}\right),
\end{equation}}where $\mu$ and $\sigma$ denote the location and scale parameters.

\subsubsection{Training Sampling as a Stochastic Process with Logit-Normal Marginals}

At each step, we sample one noise level for every frame $t$ in the training window.
This noise level sequence constitutes a stochastic process with Logit-normal marginals, parameterized by $\theta_t=(\mu_t,\sigma_t,\rho)$, where

\begin{itemize}
    \item $\mu_t$ and $\sigma_t$: the location parameter and the scale parameter in logit space at frame $t$ that specifies mean and standard deviation of the underlying Gaussian distribution.
    \item \textbf{$\rho$}: the inter-frame correlation parameter that characterizes the temporal dependence of the stochastic process.
\end{itemize}

Within this parameterized framework, the two aforementioned training strategies correspond to special cases.
Independent sampling corresponds to
\begin{equation}
\label{eq:indenpent_sample}
\mu_t = \mu_0, \quad \sigma_t = \sigma_0, \quad \rho = 0,
\end{equation}
where all frames share an identical marginal $f(k; \mu_0, \sigma_0)$, while noise levels are sampled independently across frames.
Ideal progressive sampling arises as a limiting case when
\begin{equation}
\label{eq:progressive_sample}
\mu_t \to \mathrm{logit}\Big(\frac{t}{T}\Big), \quad
\sigma_t \rightarrow 0, \quad
\rho \rightarrow 1,
\end{equation}
where the per-frame marginals in the denoising window of $T$ frames collapse to a time-ordered deterministic schedule with strong inter-frame dependence.
This formulation unifies existing paradigms and reveals the tension between noise-level coverage and training–inference consistency.

\begin{figure}[t]
  \centering
  \includegraphics[width=0.9\columnwidth]{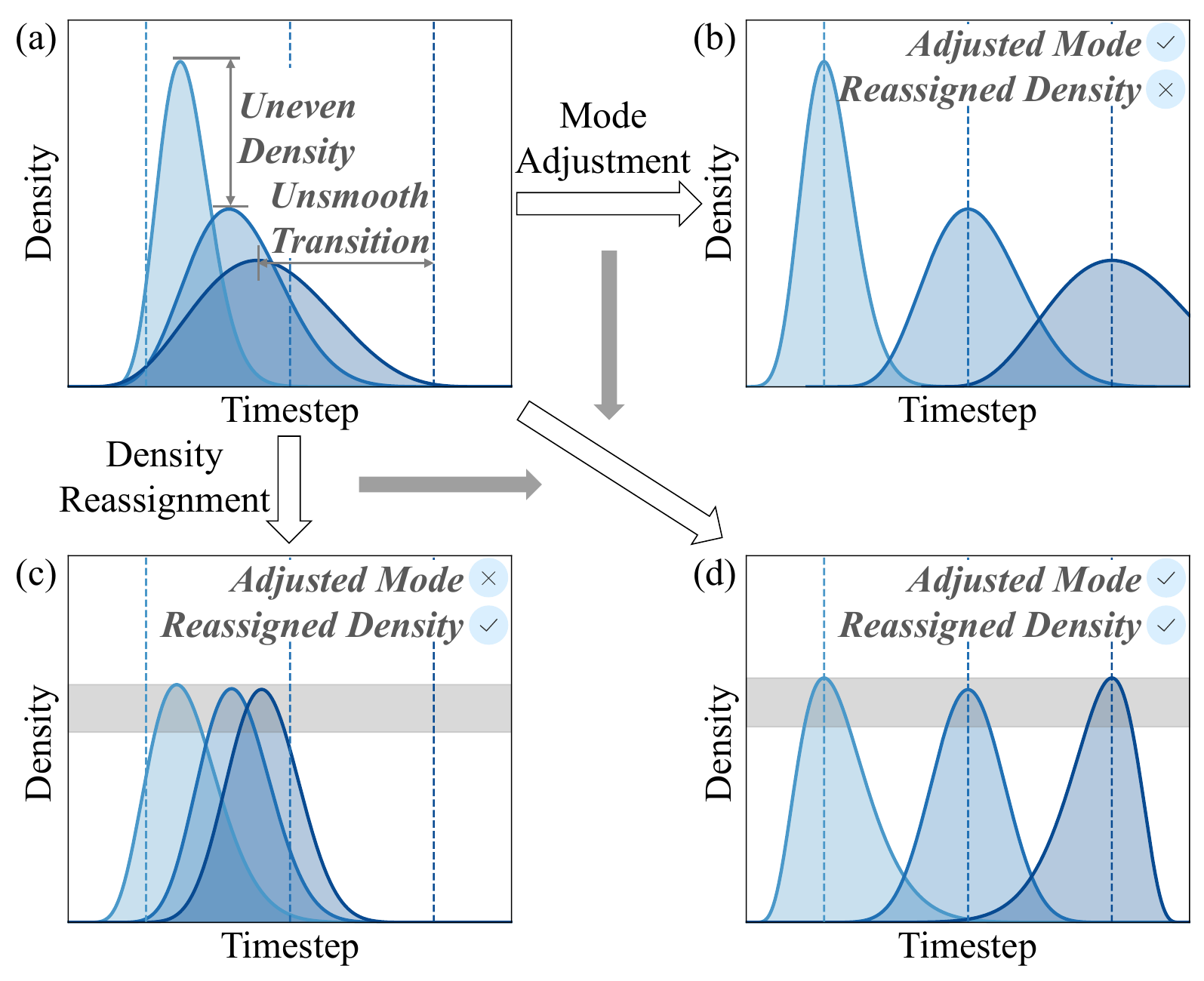}
  \caption{\textbf{Illustration of trajectory constraints.} Each subfigure shows marginals for a three-frame sampling process.
(a) Neither Constraint 1 nor Constraint 2 is satisfied.
(b) Only Constraint 1 is satisfied.
(c) Only Constraint 2 is satisfied.
(d) Both Constraints are satisfied.}
\label{fig:constraints}
\end{figure}

\begin{algorithm}[t]
\caption{Joint Calibration}
\label{alg:calibration}
\begin{algorithmic}
    \REQUIRE Initialization $(\mu_{init}, \sigma_{init})$, target modes $\{\zeta_t^*\}_{t=1}^T$
    \ENSURE Joint parameters $\{\mu_t, \sigma_t\}_{t=1}^T$
    \STATE \algbox{/*Calculate reference density*/}
    \FOR{$t = 1$ to $T$}
        \STATE Initialize $\sigma_t \gets \sigma_{init}$, $\mu_t \gets \mu_{init}$
        \STATE $h_t \gets f(\zeta_t^*;\mu_t,\sigma_t)$ \quad \algbox{/* Eq.~\ref{eq:pdf} */}
    \ENDFOR
    \STATE $H \gets \mathrm{median}(\{h_t\}_{t=1}^T)$
    \STATE
    \STATE /* Grid search for $\sigma$ */
    \FOR{$t = 1$ to $T$}
        \STATE Set $\Delta\sigma \gets 0.05$, $\Delta h \gets \infty$
        \FOR{$n = 0$ to $N-1$}
            \STATE $\hat\sigma \gets n\Delta\sigma$
            \STATE $\hat\mu \gets \psi(\zeta^*_t, \hat\sigma)$ \quad \algbox{/* Eq.~\ref{eq:modeeq} */}
            \STATE $\hat h \gets f(\zeta^*_t;\hat\mu,\hat\sigma)$ \quad \algbox{/* Eq.~\ref{eq:pdf} */}
            \IF{$\mid\mid \hat h - H \mid\mid < \Delta h$}
                \STATE Set $\sigma_t\gets \hat\sigma$, $\Delta h \gets \mid\mid \hat h - H\mid\mid$
            \ENDIF
        \ENDFOR
        \STATE Compute $\mu_t \gets \psi(\zeta_t^*, \sigma_t)$ \quad \algbox{/* Eq.~\ref{eq:modeeq} */}
    \ENDFOR
    
    \STATE \textbf{return} $\{(\mu_t, \sigma_t)\}_{t=1}^T$
\end{algorithmic}
\end{algorithm}

\subsubsection{Constrained Training Trajectory}
\label{sec:3.1.3Continuous-Trajectory}

We construct a continuous training trajectory from independent sampling to progressive sampling in the parameterized sampling-process space.
At each training step $s$, the sampling configuration is specified by $\theta^s_t=(\mu^s_t,\sigma^s_t,\rho^s)$\footnote{Throughout this paper, we adopt notational conventions that superscripts denote the training step index, while subscripts indicate the frame index within a video clip. When both are present, $x_t^s$ refers to the quantity at frame $t$ of the clip at training step $s$.}.
As training progresses, these parameters gradually evolve along the trajectory rather than remaining fixed.
A valid trajectory should provide a smooth curriculum while preserving comparable sampling coverage and appropriate temporal dependence. Accordingly, we impose the following three constraints, illustrated in Fig.~\ref{fig:constraints}.

\paragraph{Constraint 1: Smooth trajectory transition.}
Per-frame marginal distribution should evolve continuously across training steps. We therefore require the modes of the per-frame PDFs (i.e., the peak location of the PDF) to shift at an approximately uniform rate along the trajectory, avoiding abrupt changes in the sampling configuration (cf. the comparison between Fig.~\ref{fig:constraints}(b) and (a)).

\begin{figure*}[t]
  \centering
  \includegraphics[width=0.9\textwidth]{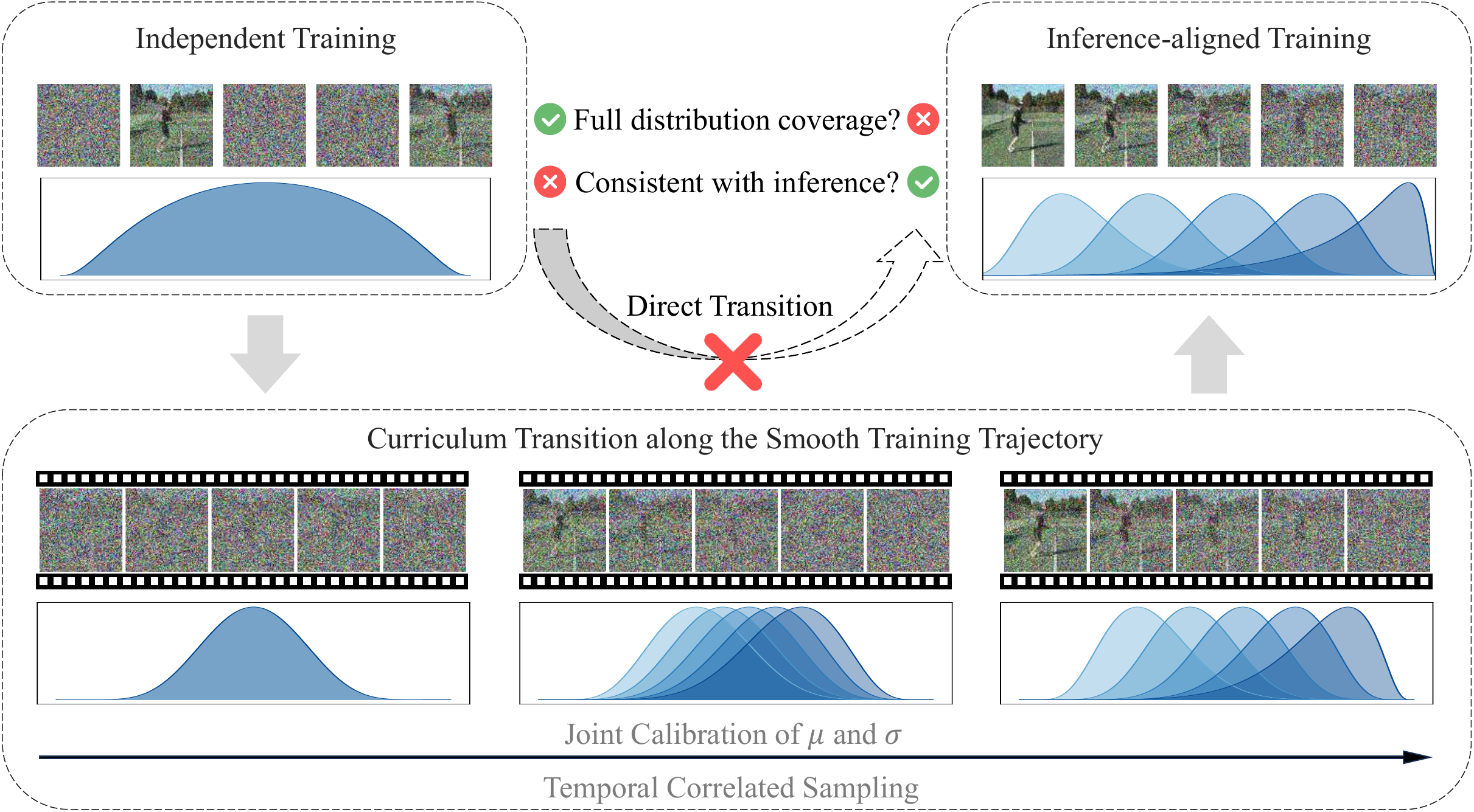} 
  \caption{\textbf{Overview of the curriculum training.} Our method builds a continuous curriculum transition between independent training and inference-aligned training that balances full distribution coverage with consistency at inference time.}
  \label{fig:main}
  \vspace{-5pt}
\end{figure*}

\paragraph{Constraint 2: Consistent per-frame coverage.}
Within each training step, the marginal distributions of different frames should exhibit comparable degrees of dispersion, so that all frames receive similarly broad coverage over noise levels.
Otherwise, some frames may be trained repeatedly within a narrow noise range, while others receive broader but sparser supervision, resulting in imbalanced per-frame optimization.
We approximate this condition by matching the densities at their modes (compare Fig.~\ref{fig:constraints}(c) and (a)).

\paragraph{Constraint 3: Inter-frame correlation.}
The correlation coefficient $\rho^s$ should also evolve as training progresses, such that the sampled noise levels preserve the  temporal dependence between successive frames throughout training.

More discussions on the motivations for these constraints can be referred to Appendix~\ref{sec:motivation}.

\subsection{Training Trajectory Implementation}
\label{sec:3.2trajectory}
We introduce a joint calibration algorithm to satisfy \textbf{Constraint 1} and \textbf{Constraint 2}, and a temporal correlative sampling algorithm to satisfy \textbf{Constraint 3}.

\subsubsection{Joint Calibration of Logit-Normal Parameters}
We optimize the two Logit-normal parameters jointly to satisfy \textbf{Constraints 1 and 2}.
Specifically, \textbf{Constraint 1} is enforced by uniformly interpolating the marginal modes between the two endpoint configurations (Eqs.~\ref{eq:indenpent_sample} \& \ref{eq:progressive_sample}).
We set $\mu_0=0$ for the initial setting in Eq.~\ref{eq:indenpent_sample}, yielding the target modes $\{\zeta_t^s\}^*$

\begin{equation}
\label{eq:curriculum}
\begin{cases}
{\zeta^s_t}^* = (1 - s)\cdot {\zeta^0_t}^* + s \cdot {\zeta^1_t}^*, \\
{\zeta^0_t}^* = 0.5, \quad
{\zeta^1_t}^* = \frac{t}{T}.
\end{cases}
\end{equation}
To ensure \textbf{Constraint 2}, we enforce a global reference peak density $H^s$ and solve the scale parameter $\sigma^s_t$ that minimize the differences between real peak densities and $H^s$.
Then, we formulate the optimization objective as
\begin{equation}
\label{eq:mode_constraint}
    \min_{\zeta^s_t,\sigma^s_t}\mid\mid \zeta^s_t - {\zeta^s_t}^*\mid\mid + \mid\mid f(\zeta^s_t; \psi(\zeta^s_t,\sigma^s_t), \sigma^s_t) - H^s\mid\mid,
\end{equation}
where $\psi(\zeta,\sigma)$ is the mode equation that determines the location parameter using the mode $\zeta$ and the scale $\sigma$:
\begin{equation}\label{eq:modeeq}
    \psi(\zeta,\sigma)=\mathrm{logit}(\zeta)+\sigma^2(1-2\zeta).
\end{equation}
We provide a proof of Eq.~\ref{eq:modeeq} in Appendix~\ref{sec:proofmodeeq}.

We introduce a joint calibration algorithm to solve the optimization problem in Eq.~\ref{eq:mode_constraint}, as detailed in Alg.~\ref{alg:calibration}.
Since all parameters are independent of the training step once the target modes are set, we omit the superscript $s$.
The joint calibration consists of two steps: estimating a global reference peak density and calibrating the marginal parameters of each frame.
First, we initialize each frame with $\mu_t=\mu_{\mathrm{init}}$ and $\sigma_t=\sigma_{\mathrm{init}}$, and compute its peak density $h_t$.
The global reference density $H$ is then defined as the median of $\{h_t\}_{t=1}^T$.
Then, we calibrate $(\mu_t,\sigma_t)$ for each target mode $\zeta_t^*$. We perform a grid search over candidate values $\hat{\sigma}$, compute the corresponding $\hat{\mu}$ using Eq.~\ref{eq:modeeq}, and select the pair whose peak density at $\zeta_t^*$ is closest to $H$. Applying this procedure at each training step yields the per-frame marginal parameters $\{\mu_t^s,\sigma_t^s\}_{t=1}^T$ satisfying Constraints 1 and 2.

\subsubsection{Temporal Correlated Sampling Strategy}

Given the calibrated per-frame marginals, we further introduce temporal correlation without altering their distributions to satisfy \textbf{Constraint 3}.
To this end, we use a Gaussian Copula-based \cite{copula} correlative sampling algorithm to decouple the temporal dependence structure from the marginal distributions and implement the coefficient $\rho^s$ gradually increasing from independent sampling with $\rho^s=0$ to strongly correlated sampling with $\rho^s\rightarrow 1$.

Specifically, for a given training step $s$, we first generate a correlated latent sequence $\{z_1, z_2, \dots, z_T\}$ in the standard normal space using a first-order autoregressive process:
\begin{equation}z_t = \rho \cdot z_{t-1} + \sqrt{1 - \rho^2} \cdot \epsilon_t,\end{equation}
where $\rho \in [0, 1]$ is the inter-frame correlation coefficient and $\epsilon_t \sim \mathcal{N}(0, 1)$.
This construction preserves the standard normal marginal of each $z_t$ while introducing temporal correlation across the sequence.
We then convert each $z_t$ into a uniform quantile using the standard normal CDF.
Finally, the quantile is mapped through the inverse CDF of the calibrated marginal distribution for frame $t$.
\begin{equation}
y_t = \mu_t + \sigma_t \cdot \Phi^{-1} \big( \Phi(z_t) \big) = \mu_t + \sigma_t \cdot z_t,
\end{equation}
where $\Phi$ presents the standard normal CDF.
The noise levels are finally generated through a sigmoid function,
\begin{equation}
k_t = \mathrm{Sigmoid}(y_t).
\end{equation}
As a result, each frame retains its prescribed marginal distribution, while the dependence induced in the Gaussian space is transferred to the sampled noise levels.

\begin{table}[t]
  \tablestyle{2.8pt}{1.1}
    \caption{16-frame unconditional generation results on UCF-101~\cite{ucf101} and Taichi-HD~\cite{taichi}.
Methods marked with * are trained on both the training and test splits, while the remaining methods are trained on the training split only.
Methods marked with \textsuperscript{\textdagger} are obtained under the experimental setting of AR-Diffusion \cite{ardiff} for a controlled comparison.}
  \label{tab:uncond-16}
  \vspace{-12pt}
  \begin{center}
        \begin{tabular}{>{\raggedright\arraybackslash}m{5.2cm}cc}
          \toprule
          Methods                         & UCF-101    &   Taichi-HD  \\

        \midrule
       
            VideoGPT* \cite{videogpt}     & 2880.6   & --      \\
            DIGAN* \cite{digan}           & 1630.2   & 156.7      \\
            StyleGAN-V*~\cite{stylegan} & 1431.0 &  143.5\\
            LVDM* \cite{lvdm}             & 372.0    & 99.0      \\
            PVDM \cite{pvdm}              & 343.6    & 267.0      \\
            FVDM \cite{fvdm}             & -- & 194.6         \\
            Latte \cite{latte}           & -- & 159.6         \\
            HVDM \cite{hvdm}              & 303.1  &  77.0      \\
            Diffusion Forcing \cite{df}   & 349.4  & 150.5        \\
            AR-Diffusion$^{\dagger}$ \cite{ardiff} & 181.9 & 100.9 \\
            
            MAGI* \cite{taming}           & 297.8  & --        \\
            FAR* \cite{far}               & 279.0  & --         \\
            FrameDiT \cite{framedit}      & -- & 95.5 \\
            \rowcolor{gray!20}
            \textbf{Ours$^{\dagger}$}                 & \textbf{146.9} & \textbf{76.1}\\
            \rowcolor{gray!20}
            \textbf{Ours}                 & \textbf{177.0} & \textbf{73.4}\\
          \bottomrule
        \end{tabular}
  \end{center}
  \vspace{-10pt}
\end{table}

\subsection{\thename{}: A Curriculum Training Procedure}
\label{sec3.3training}

Build on the above design, we propose a unified curriculum learning process that continuously evolves the training noise configuration from independent sampling toward inference-aligned progressive sampling, as illustrated in Fig.~\ref{fig:main}.
For implementation, we organize this continuous process into three successive phases: independent diffusion training, curriculum transition, and streaming-noise aligned training.

\paragraph{Independent Training.}
The training process begins with independent noise sampling to preserve broad noise-level coverage.
Specifically, the noise level of each frame is sampled independently according to Eq.~\ref{eq:indenpent_sample}.

\paragraph{Curriculum Transition.}
Following the training trajectory, we construct a smooth training trajectory transitioning from independent sampling to training-inference consistent sampling.
For practical implementation, we discretize the trajectory into uniformly-spaced points, each defining a specific training configuration, resulting in a curriculum learning schedule.
Specifically, we discretize the interval $(0, 1)$ in Eq.~\ref{eq:curriculum} into 10 uniformly-spaced configurations, and perform sequential training along this curriculum from 0 to 1.

\paragraph{Inference-Aligned Training.}
Upon the convergence of the curriculum, the training focuses exclusively on the inference-aligned noise level defined in Eq.~\ref{eq:progressive_sample}. This process aligns the model with the streamlined inference schedules.
\section{Experiments}

\begin{table}[t]
  \tablestyle{2.8pt}{1.1}
  \caption{128-frame unconditional generation results on UCF-101~\cite{ucf101} and Taichi-HD~\cite{taichi}, evaluated on the full datasets at a resolution of $256\times256$.}
  \label{tab:uncond-128}
  \vspace{-12pt}
  \begin{center}
        \begin{tabular}{>{\raggedright\arraybackslash}m{5.2cm}cc}
          \toprule
          Methods                         & UCF-101    &   Taichi-HD  \\
          \midrule
            StyleGAN-V \cite{stylegan}    & 1773.4   & 691.1      \\
            PVDM \cite{pvdm}              & 648.4    & 339.2      \\
            HVDM \cite{hvdm}              & 549.7    & 258.5       \\
            Diffusion Forcing \cite{df}   & 447.3    & 340.2      \\
            AR-Diffusion \cite{ardiff}    & 572.3 & 376.3 \\
            \rowcolor{gray!20}
            \textbf{Ours}                  & \textbf{322.5}  & \textbf{230.4} \\
          \bottomrule
        \end{tabular}
  \end{center}
  \vspace{-5pt}
\end{table}

\subsection{Comparison with Existing Baselines}

We conduct a comparison with existing baselines on UCF-101 \cite{ucf101}, and Taichi-HD \cite{taichi} at a resolution of 256×256.
For unconditional generation, we conduct short-term video generation (16 frames) as outlined in Tab.~\ref{tab:uncond-16}, and zero-shot long video extrapolation (128 frames) as outlined in Tab.~\ref{tab:uncond-128}. Results for conditional video generation and experimental settings are provided in Appendix~\ref{sec:condresults} and Appendix~\ref{sec:experimental_setup}.

\subsubsection{Video Generation Performance}

For UCF-101 \cite{ucf101}, as shown in Tab.~\ref{tab:uncond-16}, our method achieves the superior performance in unconditional video generation, with an FVD score of 177.0, improving upon the state-of-the-art by 36.6\% under the same evaluation setting. 
Similarly, on Taichi-HD \cite{taichi}, our approach achieves an FVD of 73.4, surpassing existing methods. Compared to representative methods of both progressive sampling (e.g., AR-Diffusion \cite{ardiff}) and independent sampling (e.g., Diffusion Forcing \cite{df}), our method achieves superior performance.

\subsubsection{Zero-Shot Long Video Extrapolation}
To assess the model’s capability for streaming generation, we perform zero-shot long-horizon extrapolation on 128-frame sequences, which are significantly longer than those seen during training.
Our method achieves consistently strong extrapolation performance across both datasets.
As shown in Tab.~\ref{tab:uncond-128}, our method achieves a 27.9\% and 10.9\% improvement in FVD score on UCF-101 \cite{ucf101} and Taichi-HD \cite{taichi}, respectively.
Together, these results confirm the strong zero-shot extrapolation capability of our method for long-horizon streaming generation.

\begin{table}[t]
    \tablestyle{9.3pt}{1.1}
    \caption{Comparison of driving video generation methods on the nuScenes \cite{nuscenes} test set.}
    \label{tab:drive}
    \vspace{-12pt}
    \begin{center}
        \begin{tabular}{>{\raggedright\arraybackslash}m{5.2cm}cc}
          \toprule
          Methods                         & FID    &   FVD  \\

        \midrule
       
            DriveGAN \cite{drivegan}     & 73.4   & 502.3      \\
            DriveDreamer \cite{drivedreamer}  & 52.6   & 452.0      \\
            WoVoGen \cite{wovogen} & 27.6 &  417.7 \\
            Drive-WM \cite{Drive-WM}  & 15.8    & 122.7      \\
            GenAD \cite{genad}   & 15.4    & 184.0      \\
            \rowcolor{gray!20}
            \textbf{Ours}                 & \textbf{9.4} & \textbf{105.0}\\
          \bottomrule
        \end{tabular}
  \end{center}
  \vspace{-5pt}
\end{table}

\begin{table}[t]
  \tablestyle{20pt}{1.1}
  \caption{Ablation study on the trajectory constraints. TS, DCC, and IFC denote trajectory transition smoothness (\textbf{C1}), per-frame distribution coverage consistency (\textbf{C2}), and inter-frame correlation (\textbf{C3}), respectively.}
  \label{tab:principle-ablation}
  \vspace{-12pt}
  \begin{center}
        \begin{tabular}{cccc}
          \toprule
          TS        & DCC       & IFC       & FVD            \\
          \midrule
          \ding{55} &           &           & 397.9          \\
                    & \ding{55} &           & 577.0          \\
                    &           & \ding{55} & 466.1          \\
          \ding{51} & \ding{51} & \ding{51} & \textbf{334.4} \\
          \bottomrule
        \end{tabular}
  \end{center}
\end{table}

\begin{table}[t]
  \tablestyle{21pt}{1.1}
  \caption{Ablation study on the impact of Independent Training(\textbf{IT}), Curriculum Transition(\textbf{CT}), and Inference-Aligned Training(\textbf{IAT}).}
  \label{table-stages}
  \vspace{-5pt}
        \begin{tabular}{cccc}
          \toprule
          IT    & CT    & IAT    & FVD         \\
          \midrule
          \ding{51} & \ding{55} & \ding{55} & 394.2          \\
          \ding{55} & \ding{55} & \ding{51} & 359.4          \\
          \ding{51} & \ding{55} & \ding{51} & 380.9          \\
          \ding{51} & \ding{51} & \ding{55} & 343.3          \\
          \ding{51} & \ding{51} & \ding{51} & \textbf{334.4} \\
          \bottomrule
        \end{tabular}
\end{table}

\begin{table}[t]
\centering
\begin{minipage}{0.48\linewidth}
  \tablestyle{5pt}{1.1}
  \caption{Ablation of the temporal correlation coefficient.}
  \label{table-rho}
  \vspace{-12pt}
  \begin{center}
        \begin{tabular}{lcc}
          \toprule
          $\rho$        & FVD            \\
          \midrule
          Fixed = 0.00          & 466.1          \\
          Fixed = 1.00          & 368.3          \\
          \textbf{Linear Schedule} & \textbf{334.4} \\
          \bottomrule
        \end{tabular}
  \end{center}
  \vspace{1.5em}
\end{minipage}
\hfill 
\begin{minipage}{0.5\linewidth}
  \tablestyle{16pt}{1.1}
  \caption{Ablation of the ratio between independent training and curriculum transition.}
  \label{tab:ratio12}
  \vspace{-12pt}
  \begin{center}
        \begin{tabular}{lcc}
          \toprule
          Ratio        & FVD            \\
          \midrule
          1:1          & 334.4          \\
          1:2          & 348.7          \\
          \textbf{2:1} & \textbf{318.4} \\
          3:1          & 348.8          \\
          \bottomrule
        \end{tabular}
  \end{center}
\end{minipage}
\end{table}

\subsection{Application to Autonomous Driving World Models}
To assess the applicability of our method for world modeling, we conduct experiments on the nuPlan \cite{nuplan} and nuScenes \cite{nuscenes} datasets, two widely-used large-scale autonomous driving datasets.
The model is trained on 25-frame clips at a resolution of $512\times256$ and evaluated on the nuScenes test set.
As shown in Tab.~\ref{tab:drive}, we compare with existing driving video generation methods using FID and FVD.
The results demonstrate that our training paradigm can be effectively transferred to autonomous driving scenarios, highlighting its potential as a general training strategy for driving world models.

\subsection{Ablation study}

\subsubsection{Trajectory Constraints}

We conduct ablation studies to assess the impact of the proposed design constraints, including trajectory transition smoothness (TS), per-frame distribution coverage consistency (DCC), and Inter-frame correlation (IFC), as summarized in Tab.~\ref{tab:principle-ablation}. 
Violating any individual principle leads to a notable increase in FVD. 
Among them, removing DCC causes the most severe performance degradation, underscoring the importance of maintaining smooth transitions across the training procedure.
When all three principles are jointly enforced, the model achieves the best performance, demonstrating that these principles are complementary and essential for high-quality video generation.

\begin{figure*}[t]
  \centering
  \includegraphics[width=0.93\textwidth]{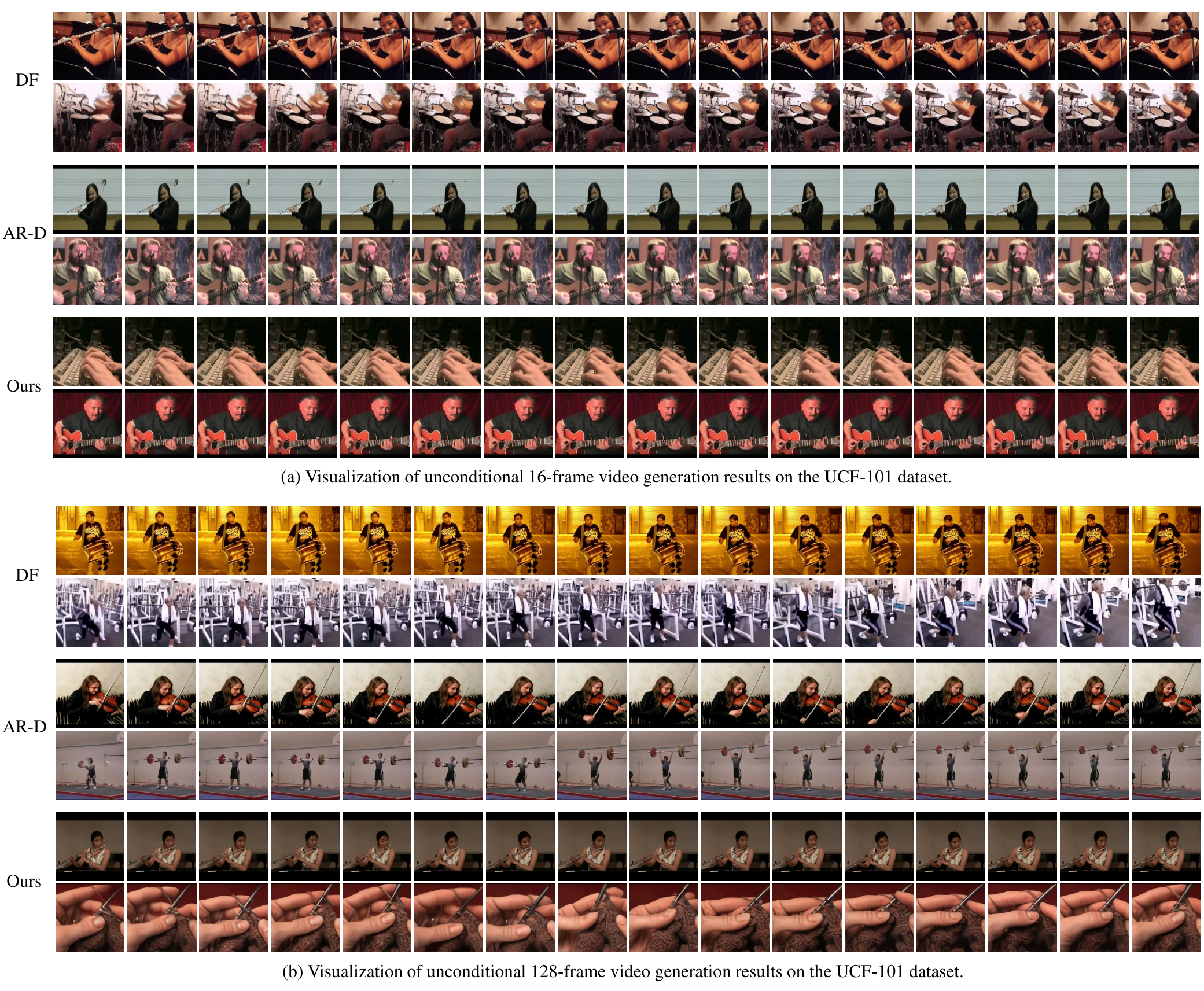} 
  \vspace{-5pt}
  \caption{\textbf{Visualization of unconditional video generation on UCF-101 dataset.} DF: Diffusion Forcing \cite{df}. AR-D: AR-Diffusion \cite{ardiff}. Our method presents higher visual quality and more robust long-horizontal extrapolation.}
  \label{fig:vis_uncond_ucf_16}
  \vspace{-12pt}
\end{figure*}

\subsubsection{Curriculum Training}
Tab.~\ref{table-stages} reports an ablation study on the contribution of our training framework.
Training only with the initial independent sampling (Row 1) yields inferior results compared to using only the inference-aligned sampling (Row 2), highlighting the importance of training–inference consistency.
Moreover, directly combining them without a smooth transition (Row 3) degrades performance.
In contrast, introducing the curriculum transition establishes a smooth training trajectory and results in performance gains (Row 4).
The complete training procedure achieves the best performance.

\subsubsection{Inter-frame correlation coefficient}
We present an ablation study on the scheduling strategy of the correlation coefficient $\rho$ in Tab.~\ref{table-rho}. Gradually increasing $\rho$ achieves the best FVD, indicating that a smooth transition from independent sampling to strongly correlated sampling better balances training diversity and temporal consistency.

\subsubsection{Ratio between independent training and curriculum transition}
We conduct an ablation study on the ratio between independent training and curriculum transition, as shown in Tab.~\ref{tab:ratio12}.
The model performs best at a 2:1 ratio, which balances training coverage with a smooth transition toward train-inference consistency.

\subsection{Qualitative Results}
As shown in Fig.~\ref{fig:vis_uncond_ucf_16}, we compare the visualization of the independent sampling method Diffusion Forcing \cite{df}, progressive sampling method AR-Diffusion \cite{ardiff}, and our method.
As illustrated in Fig.~\ref{fig:vis_uncond_ucf_16}a, Diffusion Forcing \cite{df} suffers from unstable jitter and AR-Diffusion \cite{ardiff} produces minimal motion in 16-frame generation, while our approach achieves both stable motion and high-quality details.
Fig.~\ref{fig:vis_uncond_ucf_16}b shows the visualization of 128-frame generation. Our method exhibits more coherent and realistic motion for highly dynamic actions, and remains stable in low-motion scenes.
\section{Related Work}

\subsection{Video Generation}
Video generation has recently advanced along two main paradigms: diffusion models \cite{vdm} and autoregressive architectures \cite{var,llamagen}.
Diffusion-based approaches \cite{svd,mobilei2v,vdm,latte,lvdm} perform iterative denoising to achieve high visual fidelity, while autoregressive methods \cite{videogpt,videopoet,cogvideo,loong} model videos as token sequences through causal next-token prediction.
By modeling spatiotemporal dynamics, these approaches exhibit strong potential as world models \cite{worldmodel,sora}.

\subsection{Streaming Video Generation}
Driven by the growing demand for real-time long-horizon video modeling, streaming video generation \cite{ca2vdm,acdit,artv,causalvid,ardiff,far,lct,mardini,streamingt2v} has recently attracted significant attention.
Independent sampling approaches \cite{df,dfot,fvdm,causalvid} inject independent random noise into each frame, allowing the model to be trained flexibly under diverse noise conditions.
Progressive sampling methods sample a base noise level \cite{rdm,ardiff,progressive} during training for a specific frame and construct a strictly increasing noise sequence over all video frames. 
Furthermore, self forcing \cite{sf,sf++} and rolling forcing \cite{rf} employ knowledge distillation to perform closed-loop consistent training.
These approaches are orthogonal to ours and could potentially be combined, which will be explored in future work.

\section{Conclusion}
We propose \thename{}, a unified training paradigm for streaming video generation that preserves diverse noise-level coverage while aligning training with inference.
The core idea is to construct a smooth training trajectory that connects independent and progressive sampling, allowing the model to leverage the strengths of both approaches.
To realize this, we reformulate the training sampling process within a Logit-Normal parameterized stochastic process space.
We introduce joint calibration to enforce smoothness constraints and a Gaussian Copula to enforce inter-frame correlation.
By explicitly enforcing these constraints, the training trajectory achieves a smooth transition between distribution coverage and inference alignment.
Extensive experiments demonstrate that \thename{} outperforms both single-stage and straightforward sequential training.
On two benchmark datasets, it achieves superior video generation quality and shows robust zero-shot extrapolation.
These results confirm that constructing a smooth training trajectory with the proposed constraints effectively balances distribution coverage and training–inference consistency, yielding high-quality, temporally coherent streaming video generation.

{
    \small
    \bibliographystyle{ieeenat_fullname}
    \bibliography{main}

@article{kling,
  title={Kling-Omni Technical Report},
  author={Team, Kling and Chen, Jialu and Ci, Yuanzheng and Du, Xiangyu and Feng, Zipeng and Gai, Kun and Guo, Sainan and Han, Feng and He, Jingbin and He, Kang and others},
  journal={arXiv preprint arXiv:2512.16776},
  year={2025}
}

@article{seedance,
  title={Seedance 2.0: Advancing video generation for world complexity},
  author={Seedance, Team and Chen, De and Chen, Liyang and Chen, Xin and Chen, Ying and Chen, Zhuo and Chen, Zhuowei and Cheng, Feng and Cheng, Tianheng and Cheng, Yufeng and others},
  journal={arXiv preprint arXiv:2604.14148},
  year={2026}
}

@article{wan,
  title={Wan: Open and advanced large-scale video generative models},
  author={Wan, Team and Wang, Ang and Ai, Baole and Wen, Bin and Mao, Chaojie and Xie, Chen-Wei and Chen, Di and Yu, Feiwu and Zhao, Haiming and Yang, Jianxiao and others},
  journal={arXiv preprint arXiv:2503.20314},
  year={2025}
}

@article{latte,
  title={Latte: Latent diffusion transformer for video generation},
  author={Ma, Xin and Wang, Yaohui and Chen, Xinyuan and Jia, Gengyun and Liu, Ziwei and Li, Yuan-Fang and Chen, Cunjian and Qiao, Yu},
  journal={arXiv preprint arXiv:2401.03048},
  year={2024}
}

@article{vdm,
  title={Video diffusion models},
  author={Ho, Jonathan and Salimans, Tim and Gritsenko, Alexey and Chan, William and Norouzi, Mohammad and Fleet, David J},
  journal={Advances in neural information processing systems},
  volume={35},
  pages={8633--8646},
  year={2022}
}

@article{svd,
  title={Stable video diffusion: Scaling latent video diffusion models to large datasets},
  author={Blattmann, Andreas and Dockhorn, Tim and Kulal, Sumith and Mendelevitch, Daniel and Kilian, Maciej and Lorenz, Dominik and Levi, Yam and English, Zion and Voleti, Vikram and Letts, Adam and others},
  journal={arXiv preprint arXiv:2311.15127},
  year={2023}
}

@article{ca2vdm,
  title={Ca2-vdm: Efficient autoregressive video diffusion model with causal generation and cache sharing},
  author={Gao, Kaifeng and Shi, Jiaxin and Zhang, Hanwang and Wang, Chunping and Xiao, Jun and Chen, Long},
  journal={arXiv preprint arXiv:2411.16375},
  year={2024}
}

@article{df,
  title={Diffusion forcing: Next-token prediction meets full-sequence diffusion},
  author={Chen, Boyuan and Mart{\'\i} Mons{\'o}, Diego and Du, Yilun and Simchowitz, Max and Tedrake, Russ and Sitzmann, Vincent},
  journal={Advances in Neural Information Processing Systems},
  volume={37},
  pages={24081--24125},
  year={2024}
}

@article{dfot,
  title={History-guided video diffusion},
  author={Song, Kiwhan and Chen, Boyuan and Simchowitz, Max and Du, Yilun and Tedrake, Russ and Sitzmann, Vincent},
  journal={arXiv preprint arXiv:2502.06764},
  year={2025}
}

@article{magi,
  title={MAGI-1: Autoregressive Video Generation at Scale},
  author={Teng, Hansi and Jia, Hongyu and Sun, Lei and Li, Lingzhi and Li, Maolin and Tang, Mingqiu and Han, Shuai and Zhang, Tianning and Zhang, WQ and Luo, Weifeng and others},
  journal={arXiv preprint arXiv:2505.13211},
  year={2025}
}

@inproceedings{causalvid,
  title={From slow bidirectional to fast autoregressive video diffusion models},
  author={Yin, Tianwei and Zhang, Qiang and Zhang, Richard and Freeman, William T and Durand, Fredo and Shechtman, Eli and Huang, Xun},
  booktitle={Proceedings of the Computer Vision and Pattern Recognition Conference},
  pages={22963--22974},
  year={2025}
}

@article{sf,
  title={Self Forcing: Bridging the Train-Test Gap in Autoregressive Video Diffusion},
  author={Huang, Xun and Li, Zhengqi and He, Guande and Zhou, Mingyuan and Shechtman, Eli},
  journal={arXiv preprint arXiv:2506.08009},
  year={2025}
}

@article{rf,
  title={Rolling forcing: Autoregressive long video diffusion in real time},
  author={Liu, Kunhao and Hu, Wenbo and Xu, Jiale and Shan, Ying and Lu, Shijian},
  journal={arXiv preprint arXiv:2509.25161},
  year={2025}
}

@article{mobilei2v,
  title={MobileI2V: Fast and High-Resolution Image-to-Video on Mobile Devices},
  author={Zhang, Shuai and Tang, Bao and Yu, Siyuan and Zhu, Yueting and Yao, Jingfeng and Zou, Ya and Yuan, Shanglin and Yu, Li and Liu, Wenyu and Wang, Xinggang},
  journal={arXiv preprint arXiv:2511.21475},
  year={2025}
}

@article{videogpt,
  title={Videogpt: Video generation using vq-vae and transformers},
  author={Yan, Wilson and Zhang, Yunzhi and Abbeel, Pieter and Srinivas, Aravind},
  journal={arXiv preprint arXiv:2104.10157},
  year={2021}
}

@article{cogvideo,
  title={Cogvideo: Large-scale pretraining for text-to-video generation via transformers},
  author={Hong, Wenyi and Ding, Ming and Zheng, Wendi and Liu, Xinghan and Tang, Jie},
  journal={arXiv preprint arXiv:2205.15868},
  year={2022}
}

@inproceedings{videopoet,
  title={VideoPoet: A Large Language Model for Zero-Shot Video Generation},
  author={Kondratyuk, Dan and Yu, Lijun and Gu, Xiuye and Lezama, Jose and Huang, Jonathan and Schindler, Grant and Hornung, Rachel and Birodkar, Vighnesh and Yan, Jimmy and Chiu, Ming-Chang and others},
  booktitle={International Conference on Machine Learning},
  pages={25105--25124},
  year={2024},
  organization={PMLR}
}

@article{loong,
  title={Loong: Generating minute-level long videos with autoregressive language models},
  author={Wang, Yuqing and Xiong, Tianwei and Zhou, Daquan and Lin, Zhijie and Zhao, Yang and Kang, Bingyi and Feng, Jiashi and Liu, Xihui},
  journal={arXiv preprint arXiv:2410.02757},
  year={2024}
}

@article{acdit,
  title={Acdit: Interpolating autoregressive conditional modeling and diffusion transformer},
  author={Hu, Jinyi and Hu, Shengding and Song, Yuxuan and Huang, Yufei and Wang, Mingxuan and Zhou, Hao and Liu, Zhiyuan and Ma, Wei-Ying and Sun, Maosong},
  journal={arXiv preprint arXiv:2412.07720},
  year={2024}
}

@inproceedings{artv,
  title={Art-v: Auto-regressive text-to-video generation with diffusion models},
  author={Weng, Wenming and Feng, Ruoyu and Wang, Yanhui and Dai, Qi and Wang, Chunyu and Yin, Dacheng and Zhao, Zhiyuan and Qiu, Kai and Bao, Jianmin and Yuan, Yuhui and others},
  booktitle={Proceedings of the IEEE/CVF Conference on Computer Vision and Pattern Recognition},
  pages={7395--7405},
  year={2024}
}

@inproceedings{ardiff,
  title={Ar-diffusion: Asynchronous video generation with auto-regressive diffusion},
  author={Sun, Mingzhen and Wang, Weining and Li, Gen and Liu, Jiawei and Sun, Jiahui and Feng, Wanquan and Lao, Shanshan and Zhou, SiYu and He, Qian and Liu, Jing},
  booktitle={Proceedings of the Computer Vision and Pattern Recognition Conference},
  pages={7364--7373},
  year={2025}
}

@article{far,
  title={Long-context autoregressive video modeling with next-frame prediction},
  author={Gu, Yuchao and Mao, Weijia and Shou, Mike Zheng},
  journal={arXiv preprint arXiv:2503.19325},
  year={2025}
}

@article{lct,
  title={Long context tuning for video generation},
  author={Guo, Yuwei and Yang, Ceyuan and Yang, Ziyan and Ma, Zhibei and Lin, Zhijie and Yang, Zhenheng and Lin, Dahua and Jiang, Lu},
  journal={arXiv preprint arXiv:2503.10589},
  year={2025}
}

@article{mardini,
  title={Mardini: Masked autoregressive diffusion for video generation at scale},
  author={Liu, Haozhe and Liu, Shikun and Zhou, Zijian and Xu, Mengmeng and Xie, Yanping and Han, Xiao and P{\'e}rez, Juan C and Liu, Ding and Kahatapitiya, Kumara and Jia, Menglin and others},
  journal={arXiv preprint arXiv:2410.20280},
  year={2024}
}

@inproceedings{streamingt2v,
  title={Streamingt2v: Consistent, dynamic, and extendable long video generation from text},
  author={Henschel, Roberto and Khachatryan, Levon and Poghosyan, Hayk and Hayrapetyan, Daniil and Tadevosyan, Vahram and Wang, Zhangyang and Navasardyan, Shant and Shi, Humphrey},
  booktitle={Proceedings of the Computer Vision and Pattern Recognition Conference},
  pages={2568--2577},
  year={2025}
}

@article{sf++,
  title={Self-Forcing++: Towards Minute-Scale High-Quality Video Generation},
  author={Cui, Justin and Wu, Jie and Li, Ming and Yang, Tao and Li, Xiaojie and Wang, Rui and Bai, Andrew and Ban, Yuanhao and Hsieh, Cho-Jui},
  journal={arXiv preprint arXiv:2510.02283},
  year={2025}
}

@article{fvdm,
  title={Redefining temporal modeling in video diffusion: The vectorized timestep approach},
  author={Liu, Yaofang and Ren, Yumeng and Cun, Xiaodong and Artola, Aitor and Liu, Yang and Zeng, Tieyong and Chan, Raymond H and Morel, Jean-michel},
  journal={arXiv preprint arXiv:2410.03160},
  year={2024}
}

@article{fifo,
  title={Fifo-diffusion: Generating infinite videos from text without training},
  author={Kim, Jihwan and Kang, Junoh and Choi, Jinyoung and Han, Bohyung},
  journal={Advances in Neural Information Processing Systems},
  volume={37},
  pages={89834--89868},
  year={2024}
}

@inproceedings{rdm,
  title={Rolling diffusion models},
  author={Ruhe, David and Heek, Jonathan and Salimans, Tim and Hoogeboom, Emiel},
  booktitle={Proceedings of the International Conference on Machine Learning (ICML)},
  year={2024}
}

@inproceedings{progressive,
  title={Progressive autoregressive video diffusion models},
  author={Xie, Desai and Xu, Zhan and Hong, Yicong and Tan, Hao and Liu, Difan and Liu, Feng and Kaufman, Arie and Zhou, Yang},
  booktitle={Proceedings of the Computer Vision and Pattern Recognition Conference},
  pages={6322--6332},
  year={2025}
}

@article{digan,
  title={Generating videos with dynamics-aware implicit generative adversarial networks},
  author={Yu, Sihyun and Tack, Jihoon and Mo, Sangwoo and Kim, Hyunsu and Kim, Junho and Ha, Jung-Woo and Shin, Jinwoo},
  journal={arXiv preprint arXiv:2202.10571},
  year={2022}
}

@inproceedings{stylegan,
  title={Stylegan-v: A continuous video generator with the price, image quality and perks of stylegan2},
  author={Skorokhodov, Ivan and Tulyakov, Sergey and Elhoseiny, Mohamed},
  booktitle={Proceedings of the IEEE/CVF conference on computer vision and pattern recognition},
  pages={3626--3636},
  year={2022}
}

@inproceedings{pvdm,
  title={Video probabilistic diffusion models in projected latent space},
  author={Yu, Sihyun and Sohn, Kihyuk and Kim, Subin and Shin, Jinwoo},
  booktitle={Proceedings of the IEEE/CVF conference on computer vision and pattern recognition},
  pages={18456--18466},
  year={2023}
}

@inproceedings{taming,
  title={Taming teacher forcing for masked autoregressive video generation},
  author={Zhou, Deyu and Sun, Quan and Peng, Yuang and Yan, Kun and Dong, Runpei and Wang, Duomin and Ge, Zheng and Duan, Nan and Zhang, Xiangyu},
  booktitle={Proceedings of the Computer Vision and Pattern Recognition Conference},
  pages={7374--7384},
  year={2025}
}

@article{lvdm,
  title={Latent video diffusion models for high-fidelity long video generation},
  author={He, Yingqing and Yang, Tianyu and Zhang, Yong and Shan, Ying and Chen, Qifeng},
  journal={arXiv preprint arXiv:2211.13221},
  year={2022}
}

@article{Neural-RDM,
  title={Neural residual diffusion models for deep scalable vision generation},
  author={Ma, Zhiyuan and Zhao, Liangliang and Qi, Biqing and Zhou, Bowen},
  journal={Advances in Neural Information Processing Systems},
  volume={37},
  pages={117456--117480},
  year={2024}
}

@article{worldmodel,
  title     = {World models},
  author    = {Ha, David and Schmidhuber, J{\"u}rgen},
  journal   = {arXiv preprint arXiv:1803.10122},
  volume    = {2},
  number    = {3},
  year      = {2018}
}

@article{sora,
  title     = {Video generation models as world simulators},
  author    = {Tim Brooks and Bill Peebles and Connor Holmes and Will DePue and Yufei Guo and Li Jing and David Schnurr and Joe Taylor and Troy Luhman and Eric Luhman and Clarence Ng and Ricky Wang and Aditya Ramesh},
  year      = {2024}
}

@inproceedings{chi2025empowering,
  title     = {Empowering World Models with Reflection for Embodied Video Prediction},
  author    = {Xiaowei Chi and Chun-Kai Fan and Hengyuan Zhang and Xingqun Qi and Rongyu Zhang and Anthony Chen and Chi-Min Chan and Wei Xue and Qifeng Liu and Shanghang Zhang and Yike Guo},
  booktitle = {Forty-second International Conference on Machine Learning},
  year      = {2025}
}

@inproceedings{worldsimbench,
  title     = {WorldSimBench: Towards Video Generation Models as World Simulators},
  author    = {Yiran Qin and Zhelun Shi and Jiwen Yu and Xijun Wang and Enshen Zhou and Lijun Li and Zhenfei Yin and Xihui Liu and Lu Sheng and Jing Shao and LEI BAI and Ruimao Zhang},
  booktitle = {Forty-second International Conference on Machine Learning},
  year      = {2025}
}

@article{drivelaw,
  title     = {DriveLaW: Unifying Planning and Video Generation in a Latent Driving World},
  author    = {Xia, Tianze and Li, Yongkang and Zhou, Lijun and Yao, Jingfeng and Xiong, Kaixin and Sun, Haiyang and Wang, Bing and Ma, Kun and Ye, Hangjun and Liu, Wenyu and others},
  journal   = {arXiv preprint arXiv:2512.23421},
  year      = {2025}
}

@article{rad,
  title={Rad-2: Scaling reinforcement learning in a generator-discriminator framework},
  author={Gao, Hao and Chen, Shaoyu and Zhu, Yifan and Song, Yuehao and Liu, Wenyu and Zhang, Qian and Wang, Xinggang},
  journal={arXiv preprint arXiv:2604.15308},
  year={2026}
}

@inproceedings{hvdm,
  title={Hybrid video diffusion models with 2d triplane and 3d wavelet representation},
  author={Kim, Kihong and Lee, Haneol and Park, Jihye and Kim, Seyeon and Lee, Kwanghee and Kim, Seungryong and Yoo, Jaejun},
  booktitle={European Conference on Computer Vision},
  pages={148--165},
  year={2024},
  organization={Springer}
}

@article{scalingnoise,
  title={Scalingnoise: Scaling inference-time search for generating infinite videos},
  author={Yang, Haolin and Tang, Feilong and Hu, Ming and Yin, Qingyu and Li, Yulong and Liu, Yexin and Peng, Zelin and Gao, Peng and He, Junjun and Ge, Zongyuan and others},
  journal={arXiv preprint arXiv:2503.16400},
  year={2025}
}

@article{ucf101,
  title={Ucf101: A dataset of 101 human actions classes from videos in the wild},
  author={Soomro, Khurram and Zamir, Amir Roshan and Shah, Mubarak},
  journal={arXiv preprint arXiv:1212.0402},
  year={2012}
}

@article{taichi,
  title={First order motion model for image animation},
  author={Siarohin, Aliaksandr and Lathuili{\`e}re, St{\'e}phane and Tulyakov, Sergey and Ricci, Elisa and Sebe, Nicu},
  journal={Advances in neural information processing systems},
  volume={32},
  year={2019}
}

@article{logitnorm,
  title={Logistic-normal distributions: Some properties and uses},
  author={Atchison, Jhon and Shen, Sheng M},
  journal={Biometrika},
  volume={67},
  number={2},
  pages={261--272},
  year={1980},
  publisher={Oxford University Press}
}

@inproceedings{sd3,
  title={Scaling rectified flow transformers for high-resolution image synthesis},
  author={Esser, Patrick and Kulal, Sumith and Blattmann, Andreas and Entezari, Rahim and M{\"u}ller, Jonas and Saini, Harry and Levi, Yam and Lorenz, Dominik and Sauer, Axel and Boesel, Frederic and others},
  booktitle={Forty-first international conference on machine learning},
  year={2024}
}

@article{llamagen,
  title={Autoregressive Model Beats Diffusion: Llama for Scalable Image Generation},
  author={Sun, Peize and Jiang, Yi and Chen, Shoufa and Zhang, Shilong and Peng, Bingyue and Luo, Ping and Yuan, Zehuan},
  journal={arXiv preprint arXiv:2406.06525},
  year={2024}
}

@article{var,
  title={Visual autoregressive modeling: Scalable image generation via next-scale prediction},
  author={Tian, Keyu and Jiang, Yi and Yuan, Zehuan and Peng, Bingyue and Wang, Liwei},
  journal={Advances in neural information processing systems},
  volume={37},
  pages={84839--84865},
  year={2024}
}

@InBook{copula,
    author      = {Nelsen, R. B.},
    title       = {An Introduction to Copulas},
    address     = {New York},
    edition     = {2nd},
    publisher   = {Springer-Verlag},
    year        = {2006}
}

@article{arvae,
  title={An image is worth 32 tokens for reconstruction and generation},
  author={Yu, Qihang and Weber, Mark and Deng, Xueqing and Shen, Xiaohui and Cremers, Daniel and Chen, Liang-Chieh},
  journal={Advances in Neural Information Processing Systems},
  volume={37},
  pages={128940--128966},
  year={2024}
}

@article{fvd,
  title={Towards accurate generative models of video: A new metric \& challenges},
  author={Unterthiner, Thomas and Van Steenkiste, Sjoerd and Kurach, Karol and Marinier, Raphael and Michalski, Marcin and Gelly, Sylvain},
  journal={arXiv preprint arXiv:1812.01717},
  year={2018}
}

@article{pfp,
  title={Pretraining Frame Preservation in Autoregressive Video Memory Compression},
  author={Zhang, Lvmin and Cai, Shengqu and Li, Muyang and Zeng, Chong and Lu, Beijia and Rao, Anyi and Han, Song and Wetzstein, Gordon and Agrawala, Maneesh},
  journal={arXiv preprint arXiv:2512.23851},
  year={2025}
}

@article{framepack,
  title={Packing input frame context in next-frame prediction models for video generation},
  author={Zhang, Lvmin and Agrawala, Maneesh},
  journal={arXiv preprint arXiv:2504.12626},
  year={2025}
}

@article{framedit,
  title={FrameDiT: Diffusion Transformer with Matrix Attention for Efficient Video Generation},
  author={Le, Minh Khoa and Do, Kien and Nguyen, Duc Thanh and Tran, Truyen},
  journal={arXiv preprint arXiv:2603.09721},
  year={2026}
}

@inproceedings{drivegan,
  title={Drivegan: Towards a controllable high-quality neural simulation},
  author={Kim, Seung Wook and Philion, Jonah and Torralba, Antonio and Fidler, Sanja},
  booktitle={Proceedings of the IEEE/CVF Conference on Computer Vision and Pattern Recognition},
  pages={5820--5829},
  year={2021}
}

@inproceedings{drivedreamer,
  title={Drivedreamer: Towards real-world-drive world models for autonomous driving},
  author={Wang, Xiaofeng and Zhu, Zheng and Huang, Guan and Chen, Xinze and Zhu, Jiagang and Lu, Jiwen},
  booktitle={European conference on computer vision},
  pages={55--72},
  year={2024},
  organization={Springer}
}

@inproceedings{wovogen,
  title={Wovogen: World volume-aware diffusion for controllable multi-camera driving scene generation},
  author={Lu, Jiachen and Huang, Ze and Yang, Zeyu and Zhang, Jiahui and Zhang, Li},
  booktitle={European conference on computer vision},
  pages={329--345},
  year={2024},
  organization={Springer}
}

@inproceedings{Drive-WM,
  title={Driving into the future: Multiview visual forecasting and planning with world model for autonomous driving},
  author={Wang, Yuqi and He, Jiawei and Fan, Lue and Li, Hongxin and Chen, Yuntao and Zhang, Zhaoxiang},
  booktitle={Proceedings of the IEEE/CVF Conference on Computer Vision and Pattern Recognition},
  pages={14749--14759},
  year={2024}
}

@inproceedings{genad,
  title={Generalized predictive model for autonomous driving},
  author={Yang, Jiazhi and Gao, Shenyuan and Qiu, Yihang and Chen, Li and Li, Tianyu and Dai, Bo and Chitta, Kashyap and Wu, Penghao and Zeng, Jia and Luo, Ping and others},
  booktitle={Proceedings of the IEEE/CVF Conference on Computer Vision and Pattern Recognition},
  pages={14662--14672},
  year={2024}
}

@inproceedings{nuscenes,
  title={nuscenes: A multimodal dataset for autonomous driving},
  author={Caesar, Holger and Bankiti, Varun and Lang, Alex H and Vora, Sourabh and Liong, Venice Erin and Xu, Qiang and Krishnan, Anush and Pan, Yu and Baldan, Giancarlo and Beijbom, Oscar},
  booktitle={Proceedings of the IEEE/CVF conference on computer vision and pattern recognition},
  pages={11621--11631},
  year={2020}
}

@article{nuplan,
  title={nuplan: A closed-loop ml-based planning benchmark for autonomous vehicles},
  author={Caesar, Holger and Kabzan, Juraj and Tan, Kok Seang and Fong, Whye Kit and Wolff, Eric and Lang, Alex and Fletcher, Luke and Beijbom, Oscar and Omari, Sammy},
  journal={arXiv preprint arXiv:2106.11810},
  year={2021}
}

@inproceedings{dcae,
  title={Deep compression autoencoder for efficient high-resolution diffusion models},
  author={Chen, Junyu and Cai, Han and Chen, Junsong and Xie, Enze and Yang, Shang and Tang, Haotian and Li, Muyang and Han, Song},
  booktitle={International Conference on Learning Representations},
  volume={2025},
  pages={96539--96560},
  year={2025}
}
}

\clearpage
\section*{Appendices}

\setcounter{page}{1}

\setcounter{table}{0}
\renewcommand{\thetable}{A\arabic{table}}
\renewcommand{\theHtable}{appendix.A.\arabic{table}}

\setcounter{figure}{0}
\renewcommand{\thefigure}{A\arabic{figure}}
\renewcommand{\theHfigure}{appendix.A.\arabic{figure}}

\setcounter{section}{0}
\renewcommand{\thesection}{A.\arabic{section}}
\renewcommand{\theHsection}{appendix.A.\arabic{section}}

\section{Motivation}
\label{sec:motivation}

The fundamental objective is to construct a structured transition from high-entropy independent sampling toward a more concentrated, inference-aligned sampling distribution.
Since diffusion noise levels are bounded within $(0,1)$, we parameterize their frame-wise distributions using the Logit-Normal family, which naturally respects this support and can flexibly represent distributions with varying locations, dispersions, and degrees of skewness.
To characterize and control the resulting transition, we use mode dynamics as the primary frame-wise distribution indicator and the correlation coefficient as the measure of inter-frame dependency.
We focus on the mode rather than conventional moments such as the mean because, for skewed Logit-Normal distributions, the mode directly identifies the most frequently sampled noise level.
As these high-density regions are encountered more often during training, controlling the mode provides a practical reference for regulating the curriculum trajectory.
Accordingly, Constraints 1 and 2 govern the evolution of the mode location and peak density, respectively, while Constraint 3 controls inter-frame correlation.

\begin{enumerate}
\item Trajectory transition smoothness: This constraint promotes stable optimization across training steps. Abrupt changes in the sampling distribution may introduce sudden shifts in training difficulty and destabilize the curriculum. We therefore enforce a uniform evolution of the mode location along the training trajectory, ensuring that adjacent training configurations change gradually.
\item Per-frame distribution coverage consistency: It promotes balanced optimization across frames within each training step. If the distribution of one frame is sharply concentrated while that of another is overly dispersed, their dominant noise-level regions are sampled with substantially different frequencies, resulting in uneven training emphasis across the sequence. We therefore match the mode probability densities of all frame-wise distributions to maintain comparable sampling concentration.
\item Inter-frame correlation: In the initial diffusion-forcing configuration, frame-wise noise levels are sampled independently, so their inter-frame dependencies are not explicitly modeled. In contrast, inference follows an ordered denoising process in which the noise levels of adjacent frames are strongly correlated. We thus progressively introduce inter-frame correlation along the training trajectory, explicitly aligning the joint training distribution with the sequential structure of inference.
\end{enumerate}

\section{Experimental Setup}\label{sec:experimental_setup}
\subsection{Dataset}
\textbf{UCF-101} \cite{ucf101}.
The UCF-101 dataset \cite{ucf101} is a large-scale human motion dataset that consists of 13,320 videos with 9,624 clips in the training set and 3,696 clips in the test set across 101 action classes.
We train only on the train set.

\noindent\textbf{Taichi-HD} \cite{taichi}.
The Taichi-HD dataset \cite{taichi} is a human motion dataset containing 3,103 videos with 2,818 clips in the training set and 285 clips in the test set.
We train only on the train set.

\noindent\textbf{NuPlan} \cite{nuplan}.
NuPlan is a large-scale autonomous driving dataset containing diverse real-world driving scenarios.
The training set includes 1,310 driving logs and 18,104 driving segments, covering approximately 3.49 million frames sampled at 10 Hz, with an average duration of 20 seconds.
In our experiments, we only utilize the front-view camera stream among the eight available cameras.

\noindent\textbf{NuScenes} \cite{nuscenes}.
NuScenes consists of 1,000 driving scenes with temporally consistent sensor streams and vehicle motion annotations.
The official split contains 700 training scenes, 150 validation scenes, and 150 testing scenes.
We train our model on the training split and evaluate its performance on the testing split.

\subsection{Metrics}

For experiments on the UCF-101 dataset and the Taichi-HD dataset, we evaluate the quality of generated videos by computing the Fréchet Video Distance (FVD) \cite{fvd}.
We evaluate both qualities of unconditional and conditional generation on the UCF-101 dataset, and unconditional generation on the Taichi-HD dataset. 
To validate the generalizability of our method to long-horizon tasks, we provide results with video lengths of both 16 and 128 frames.

For experiments on the nuPlan \cite{nuplan} dataset and the nuScenes \cite{nuscenes} dataset, we evaluate future video generation with 25-frame clips and measure the generation quality using Fréchet Inception Distance (FID) and Fréchet Video Distance (FVD). 

\subsection{Model}
For experiments on UCF-101 \cite{ucf101} and Taichi-HD \cite{taichi}, we use DFoT \cite{dfot} DiT as the backbone for the diffusion model, and AR-VAE introduced in AR-Diffusion \cite{ardiff}, which is a Transformer-based 1D tokenizer \cite{arvae}, producing a token sequence of length 32. The backbone DiT contains 674M parameters.

For experiments on nuScenes \cite{nuscenes}, we use the Wan 2.1 DiT \cite{wan} as the backbone for the diffusion model, and a Deep Compression Autoencoder (DCAE) \cite{dcae} fine-tuned on driving data, which produces a 16-channel latent with $8\times$ spatial and $1\times$ temporal compression. 

The detailed model configuration is provided in Tab.~\ref{tab:dit_config}.

\begin{table}[ht]
\tablestyle{14pt}{1.0}
\caption{Architecture configurations.}
\label{tab:dit_config}
\begin{tabular}{lcc}
\toprule
\textbf{Configuration} & \textbf{DFoT} & \textbf{Wan}\\
\midrule
\#Parameters & 674M & 1.3B \\
Transformer Blocks & 28 & 32 \\
Hidden Size & 1152 & 2048 \\
Attention Heads & 16 & 16 \\
Resolution & $256 \times 256$ & $512 \times 256$\\
Frames & 16 & 25 \\
VAE Channels & 4 & 16 \\
VAE Downsample & $[1,8]$ & $[1,8]$ \\
VAE Tokens / Frame & 32 & 512 \\
\bottomrule
\end{tabular}
\end{table}

\subsection{Training Details}
All experiments are conducted on 8 NVIDIA H20 GPUs. 
For experiments on UCF-101 \cite{ucf101} and Taichi-HD \cite{taichi}, training is performed with a batch size of 40 per GPU. 
The model is optimized using AdamW with $\beta_1=0.9$, $\beta_2=0.99$, and $\epsilon=10^{-8}$. 
The learning rate is set to $2\times10^{-5}$ with a 10k-step warmup. 
We adopt $v$-prediction as the training objective and use 50-step DDIM sampling with a CFG scale of 2.0 during inference. 
The maximum correlation coefficient is set to $\rho_{\max}=0.95$. 
The full training schedule consists of 600k steps, including 300k steps of independent training, 150k steps of curriculum transition, and 150k steps of inference-aligned training. 
For ablation studies, we train models for 450k steps and evaluate them on 16-frame unconditional generation on UCF-101 \cite{ucf101}. More details are provided in Tab.~\ref{tab:training_config}.
 For experiments on nuScenes \cite{nuscenes}, training is performed with a batch size of 4 per GPU. The model is optimized using AdamW with $\beta_1=0.9$, $\beta_2=0.99$, $\epsilon=10^{-8}$, and weight decay $0.01$. The learning rate is set to $5\times10^{-5}$ with a 2.5k-step cosine warmup. We adopt flow matching with velocity prediction as the training objective and use 50-step sampling with a CFG scale of $6.0$ during inference. The Copula-based scheduling uses $\rho_{\max}=0.95$. The training schedule consists of 10k steps. We evaluate 25-frame conditional video generation, using the first frame as the visual condition and generating the subsequent 24 frames.

\begin{table}[ht]
\tablestyle{23pt}{1.0}
\caption{Training and sampling configurations for experiments on UCF-101 \cite{ucf101} and Taichi-HD \cite{taichi}.}
\label{tab:training_config}
\begin{tabular}{lc}
\toprule
\textbf{Configuration} & \textbf{Value} \\
\midrule
Training Objective & $v$-prediction \\
Diffusion Timesteps & 1000 \\
Sampling Steps & 50 \\
$\rho_{\max}$ & 0.95 \\
Optimizer & AdamW \\
$\beta_1, \beta_2$ & 0.9, 0.99 \\
Learning Rate & $2\times10^{-5}$  \\
Warmup Steps & 10k \\
Gradient Clipping & 1.0 \\
Batch Size & 40 \\
EMA Decay & 0.9999 \\
CFG Scale & 2.0 \\
Independent Training & 300k steps \\
Curriculum Transition & 150k steps \\
Inference-Aligned Training & 150k steps \\
\bottomrule
\end{tabular}
\end{table}

\section{Conditional video generation}\label{sec:condresults}
We conduct conditional generation experiments on the UCF-101 dataset \cite{ucf101}.
We train the model on 16-frame clips at a resolution of $256 \times 256$.
Our method presents significant improvements over prior approaches as shown in Tab.~\ref{table-cond-UCF-101-all}.
We further evaluate the long-horizon extrapolation capability on 128-frame generation, as shown in Tab.~\ref{tab:uncond128}.
These results confirm that our method maintains strong zero-shot extrapolation capability in the conditional setting for long-horizon streaming generation.

\begin{table}[ht]
    \tablestyle{22pt}{1.0}
    \caption{Comparison on conditional video generation on UCF-101. * indicates training on train + test split, while methods without * are trained only on train. $\ddagger$ indicates results evaluated on the test split, while all other results are evaluated on the full dataset. All methods generate videos at 256×256 resolution and 16 frames.}
    \label{table-cond-UCF-101-all}
    \begin{tabular}{lc} 
        \toprule
        Methods                       &   FVD          \\
        \midrule
        Latte* \cite{latte}           & 478.0          \\
        FVDM* \cite{fvdm}             & 468.2          \\
        Neural-RDM* \cite{Neural-RDM} & 461.0          \\
        Diffusion Forcing \cite{df}   & 157.8          \\
        {Ca2-VDM$^{\ddagger}$} \cite{ca2vdm}       & 184.5          \\
        FrameDiT* \cite{framedit}       & 170.1 \\
        \rowcolor{gray!20}
        \textbf{{Ours$^{\ddagger}$}}                 & \textbf{158.3} \\
        \rowcolor{gray!20}
        \textbf{Ours}                 & \textbf{121.0} \\
        \bottomrule
    \end{tabular}
\end{table}

\begin{table}[ht]
    \tablestyle{22pt}{1.0}
    \caption{Comparison of methods on UCF-101 \cite{ucf101} for conditional generation, evaluated on the full dataset. All methods generate videos at 256×256 resolution and 128 frames.}
    \label{tab:uncond128}
    \begin{tabular}{lc}
        \toprule
        Methods & FVD  \\
        \midrule
        FIFO-Diffusion \cite{fifo}       & 596.6           \\
        Diffusion Forcing \cite{df}      & 272.3           \\
        ScalingNoise \cite{scalingnoise} & 539.2           \\
        \rowcolor{gray!20}
        \textbf{Ours}                    &  \textbf{232.4} \\
        \bottomrule
    \end{tabular}
\end{table}

\section{Proof of the Mode Equation}\label{sec:proofmodeeq}
In this section, we provide a brief proof of Eq. 9.

Assume that $X\in(0,1)$ follows a Logit-Normal distribution with parameters $(\mu,\sigma^2)$, whose density is given by
\[
f(x)
=
\frac{1}{x(1-x)\sigma\sqrt{2\pi}}
\exp\!\left(
-\frac{(logit(x)-\mu)^2}{2\sigma^2}
\right).
\]
Taking the logarithm of the density and discarding additive constants yields
\[
\ell(x)
=
-\log x
-
\log(1-x)
-
\frac{(logit(x)-\mu)^2}{2\sigma^2}.
\]
Differentiating $\ell(x)$ with respect to $x$ gives
\[
\ell'(x)
=
\frac{2x-1}{x(1-x)}
-
\frac{logit(x)-\mu}{\sigma^2\,x(1-x)}.
\]
Since $x(1-x)>0$ for all $x\in(0,1)$, the stationary points of the density satisfy
\[
(2x-1)
-
\frac{logit(x)-\mu}{\sigma^2}
=
0,
\]
which can be rearranged as
\[
logit(x)-\mu
=
\sigma^2(2x-1).
\]
If the density attains its maximum at $x=\zeta\in(0,1)$, substituting $x=\zeta$ into the above condition yields
\[
\mu
=
logit(\zeta)
+
\sigma^2(1-2\zeta).
\]
Moreover, the density vanishes as $x\to0^+$ and $x\to1^-$ and admits a unique stationary point in $(0,1)$, ensuring that this solution corresponds to the unique global maximum of the density.

\section{Why Mode-Density Matching for Constraint 2}\label{sec:c2}
Constraint 2 is implemented through a mode-based heuristic rather than a global dispersion statistic such as differential entropy. Although entropy quantifies the overall uncertainty of a distribution, it does not directly govern the sampling density near its dominant noise level. Consequently, frame-wise distributions with similar or smoothly varying entropy may still exhibit substantially different peak densities, leading to uneven training emphasis across the temporal sequence.

To examine this distinction, we replace the proposed mode-density consistency constraint with two linear entropy schedules. Specifically, the entropy is gradually decreased from $-0.2$ nats to $-7$ nats in \textit{Ent 1} and to $-3$ nats in \textit{Ent 2}, as reported in Tab.~\ref{tab:ent}. Both entropy-based variants yield higher FVD than our mode-based design. These results indicate that controlling global distributional uncertainty alone is insufficient. Instead, directly matching the dominant sampling density across frames provides a more effective mechanism for balancing training emphasis and improving temporal generation quality.

\begin{table}[ht]
    \centering
    \tablestyle{48pt}{1.0}
    \caption{Comparison between entropy and mode DCC.}
    \label{tab:ent}
    \begin{tabular}{lc}
    \toprule
    Setting &	FVD    \\
    \midrule
    Ent 1   &  675 \\
    Ent 2  &    590 \\
    \rowcolor{gray!20}
    \textbf{Ours}   &  \textbf{334} \\
    \bottomrule
    \end{tabular}
\end{table}

\section{Limitation and Future Work}\label{sec:limitation}

\subsection{Limitation}
Due to computational and data constraints, experiments are conducted on small- to medium-scale datasets.
While sufficient to demonstrate the effectiveness and robustness of \thename{}, this setting does not fully reflect its behavior under large-scale training. We leave a systematic study of scaling to future work.

\subsection{Future Work}
Several orthogonal directions, such as memory compression techniques \cite{pfp,framepack} and closed-loop distillation methods \cite{sf,sf++,rf}, remain to be jointly explored, and their interactions with the proposed framework have not yet been systematically studied.
Extending the method to large-scale training on top of pretrained models \cite{magi} is a promising direction.

\section{More Visualization}\label{sec:morevis}

\begin{figure*}[t]
  \centering
  \includegraphics[width=\textwidth]{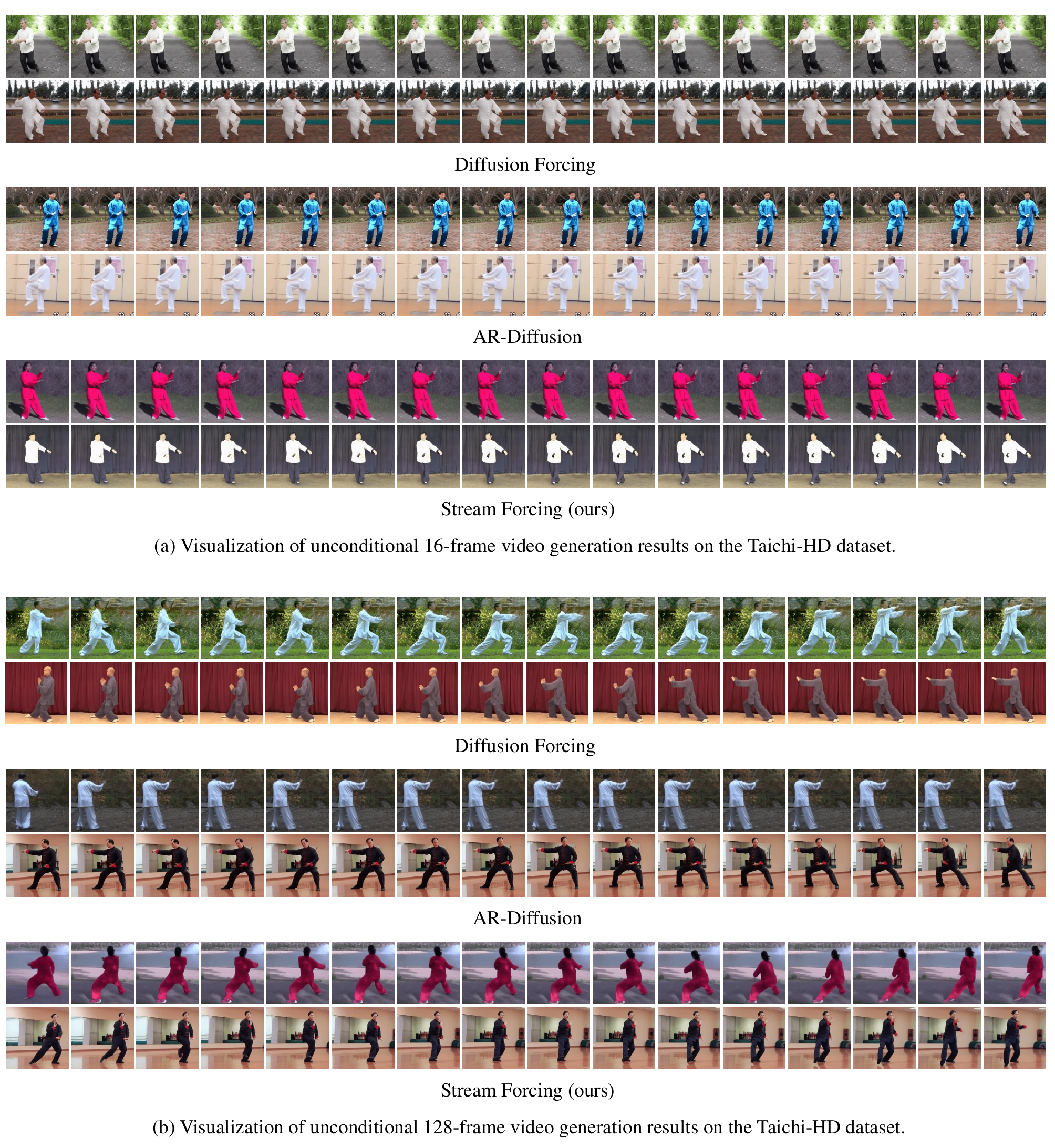} 
  \caption{\textbf{Visualization of unconditional video generation on Taichi-HD dataset.}}
  \label{fig:vis_taichi}
\end{figure*}

\begin{figure*}[t]
  \centering
  \includegraphics[width=\textwidth]{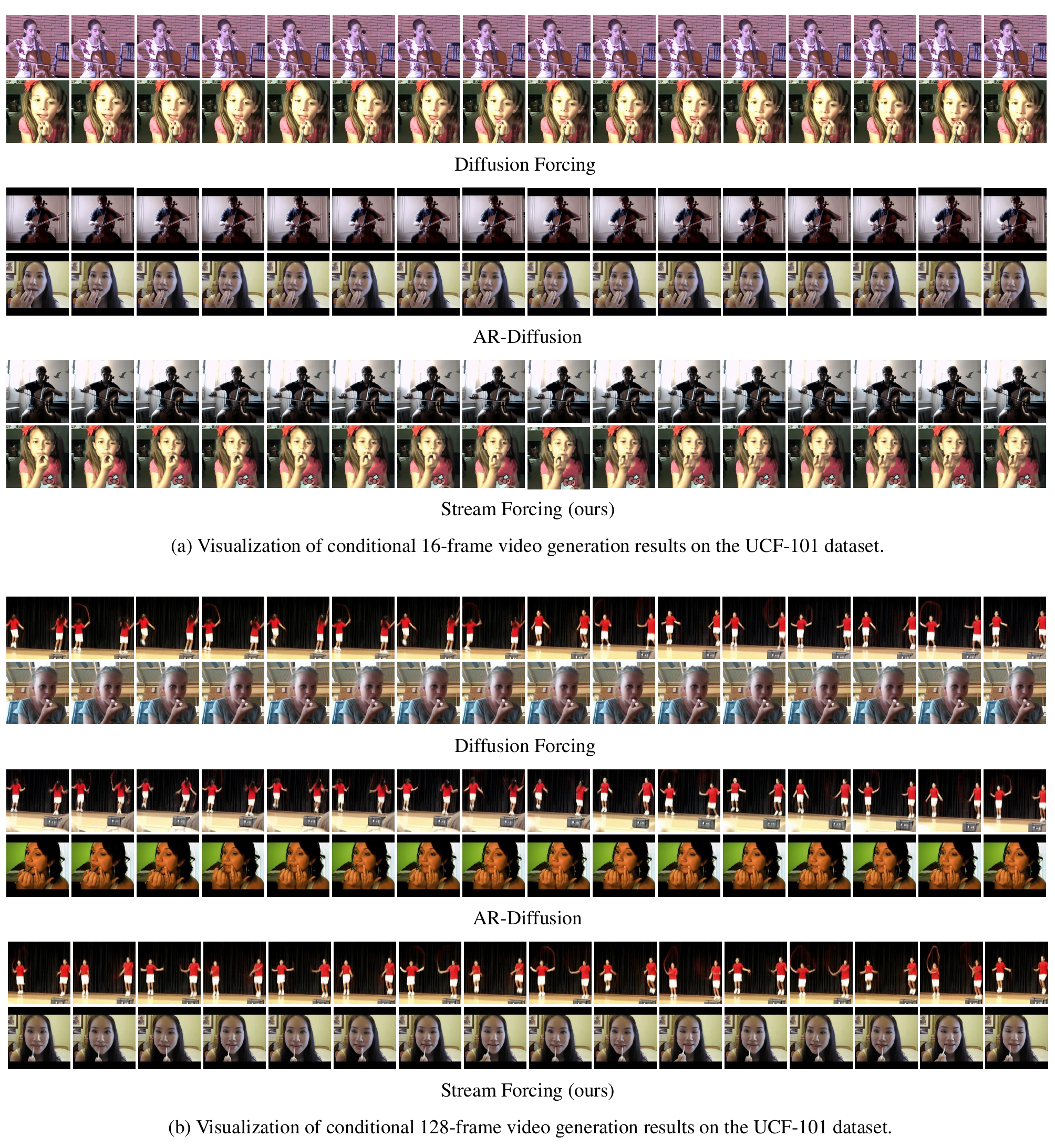} 
  \caption{\textbf{Visualization of conditional video generation on UCF-101 dataset.}}
  \label{fig:vis_cond}
\end{figure*}

In this section, we present additional visualizations of our method.
Specifically, we show unconditional video generation on the Taichi-HD dataset \cite{taichi} for 16-frame sequences (Fig.~\ref{fig:vis_taichi}a) and long-horizon extrapolation to 128 frames (Fig.~\ref{fig:vis_taichi}b).
We also show conditional video generation on the UCF-101 dataset \cite{ucf101} for 16-frame sequences Fig.~\ref{fig:vis_cond}a) and extrapolation to 128 frames (Fig.~\ref{fig:vis_cond}a).
These results demonstrate that our method maintains temporal consistency and generates visually coherent sequences across both datasets and time horizons.

\end{document}